\documentclass{article}

\usepackage[sort, numbers]{natbib}
\usepackage{float}
\usepackage{authblk}
\usepackage{amsmath}
\usepackage{bbm}
\usepackage{placeins}
\usepackage{arxiv}

\usepackage[utf8]{inputenc} 
\usepackage[T1]{fontenc}    
\usepackage{hyperref}       
\usepackage{url}            
\usepackage{booktabs}       
\usepackage{amsfonts}       
\usepackage{nicefrac}       

\usepackage{microtype}      

\usepackage{lipsum}
\usepackage{graphicx}
\graphicspath{ {./images/} }

\title{Towards Better Shale Gas Production Forecasting Using Transfer Learning}

\author[1,2,*]{Omar S. Alolayan}
\author[1,2]{Samuel J. Raymond}
\author[1,2,3]{Justin  B. Montgomery}
\author[1,2]{John R. Williams}
\affil[1]{\footnotesize Department of Civil and Environmental Engineering, Massachusetts Institute of Technology, Cambridge, MA 02139.}
\affil[2]{\footnotesize The Center for Computational Science and Engineering, Massachusetts Institute of Technology, Cambridge, MA 02139.}
\affil[3]{\footnotesize MIT Energy Initiative, Massachusetts Institute of Technology, Cambridge, MA 02139.}
\affil[*]{\footnotesize Corresponding author: Omar S. Alolayan, olyanos@mit.edu}

\begin{document}
\maketitle
\begin{abstract}
Deep neural networks can generate more accurate shale gas production forecasts in counties with a limited number of sample wells by utilizing transfer learning. This paper provides a way of transferring the knowledge gained from other deep neural network models trained on adjacent counties into the county of interest. The paper uses data from more than 6000 shale gas wells across 17 counties from Texas Barnett and Pennsylvania Marcellus shale formations to test the capabilities of transfer learning. The results reduce the forecasting error between 11\% and 47\% compared to the widely used Arps decline curve model.
\end{abstract}

\section{Introduction}
Thanks to hydraulic fracturing technology, extracting oil and gas in an unconventional shale formation became economically viable making the United States the top oil and gas producer in the world in 2020 according to the US Energy Information Administration (EIA) \citep{EIA_website}. Accurate production forecasting is necessary for many reasons such as decision making, developing returns on investments, and maintaining and managing the wells. As natural gas consumption is expected to grow in the near future increasing the demand for such resource \citep{McKinsey_2021}, developing accurate production forecasting models plays an important role in helping policy makers better assess the supply and demand situation and allocate resources based on better understanding of the matter.\\
The most accurate method for oil and gas production forecasting is reservoir simulation. However, current reservoir simulation techniques that are used in forecasting conventional oil and gas wells still face some challenges when it comes to forecasting unconventional resources due to challenges in modeling \citep{Sidel}, costly data acquisition \citep{Clarkson}, complex fracture network \citep{Lee} and variation of fracture permeability \citep{Han}.
Another approach used to forecast unconventional gas resources is decline curve analysis (DCA) in which different models have been developed and used over the years \citep{Lei,Manda}. Some of these decline curves were developed from physical interpretation of gas flow \citep{Patzek,Deholanda} and some of them are developed empirically based on observations \citep{Duong,Clark}.\\
One of the most common ones that is also used by EIA is the hyperbolic function known as the Arps model \citep{Arps}.
Arps decline curves are cheap,fast and easy to implement, and the model can generate forecasts with a minimal amount of data. However, it is a heuristic approach based on empirical observation and does not generate accurate predictions for unconventional gas wells as it tends to overestimate the production after the early stages of a well's production \citep{Qun_2020}. \citet{Justin} show that DCA shale gas forecasting is an ill-posed problem due to non-unique model parameter estimates especially when the production history is short.\\
The availability of historical data along with the recent advancement in machine learning granted a new method for forecasting production by using a deep neural network (DNN) that helps generate a more accurate production forecast. \citet{Alfattah} developed a DNN to forecast the US cumulative natural gas production using different physical and economic input parameters such as the 
gas annual depletion rate and growth rate gross domestic product (GDP). DNNs also have been used in order to improve decline curves where
\citet{Li} used a DNN to get an estimation of parameters in a logistic decline curve model. \citet{Sun_2018} used recurrent neural networks to build time series models that forecast the production using the well's production history and the tubing head pressure as input.\\
\citet{Justin} developed a DNN forecasting model that used data aggregated from multiple counties in Texas Barnett shale to forecast production in which the first few months of actual production data is used to forecast the production for the next few years. This paper uses a similar approach in which transfer learning is used to build county-specific DNN models by utilizing the knowledge acquired from other nearby counties.
In general, the DNN can be expressed as a function $\mathbbmss{F}$ that is theoretically able to learn the best non-linear mapping from the input data, which are the first few months of production, to the output data which are the next few years of production  \citep{Justin}.
A DNN can be trained by dividing the well production data into input data and label (output) data, the input data are the first 4,6,8 or 10 cumulative monthly production while the label data are the remaining cumulative monthly production data. The function $\mathbbmss{F}$ is then trained to map a specific well's first few months of cumulative production into a forecast of its future production.\\
This paper presents a novel approach in which transfer learning DNN  models can reduce the error in forecasting between 11\% and 47\%  compared to Arps decline curve models. This paper also demonstrates that by utilizing transfer learning to apply the knowledge acquired from production data in nearby counties into counties with limited data, this approach helps mitigate a major disadvantage that machine learning models have which is the need for a large amount of data to develop a good predictive model. In addition, this paper also shows how transfer learning can still be useful in generating accurate production forecasts even when the data available for a certain county is not enough to properly train a transfer learning model.\\
The paper starts first by going over the details data sets used for this study in Section \ref{section:Data}. Then, in Section \ref{section:Methods}, it explains the methodology in which the Arps decline curve model is used as a benchmark for the suggested new approach and the details of the comparison procedure between the two models. Next, in Section \ref{section:Methods} and \ref{ResultsSection}, this paper shows that DNNs can map input production data into output future production forecast and hence can be used effectively to develop shale gas production forecasts. After that, Section \ref{ResultsSection} shows how the technique of transfer learning is utilized to overcome data availability that limits DNNs from developing county-specific models. Lastly, Section \ref{Discussion} offers a thorough discussion and analysis of the results and its potential for further improvements in the future.

\section{Data}\label{section:Data}
Two data sets were used in this research paper. The first data set is composed of 4439 wells from Texas Barnett shale with 120 months worth of cumulative monthly production data \citep{TX_website}. The second data set is composed of 2172 gas wells from the Marcellus shale in Pennsylvania 
with 66 months worth of cumulative monthly production data \citep{PA_website}. The cumulative production in both data sets is
measured in thousands of standard cubic feet (Mscf). In addition, both data sets contain which county the well is located in. A sample well from the data is shown in Table 1 and Table 2 where the Well-API refers to well-specific unique identifier. 
\begin{table}[H] 
\centering

\small
\begin{tabular}{|p{2.25cm}|p{1.75cm}|p{1.75cm}|p{1.5cm}|p{0.5cm}|p{2cm}|}
 \hline
 \multicolumn{6}{|c|}{Table 1: Sample well from Texas Barnett shale} \\
 \hline
Well-API & County &State & Month-1 & ... &Month-120\\
 \hline
42-425-30160& Somervell & Texas &21295 &...& 1373\\

 \hline
\end{tabular}
\end{table}

\noindent 
\begin{table}[H]
\centering

\small
\begin{tabular}{|p{2.25cm}|p{1.75cm}|p{1.75cm}|p{1.5cm}|p{0.5cm}|p{2cm}|}
 \hline
 \multicolumn{6}{|c|}{Table 2: Sample well from Pennsylvania Marcellus shale} \\
 \hline
Well-API & County &State & Month-1 & ... &Month-66\\
 \hline
059-26033& Greene & Pennsylvania &87601 &...& 6310\\

 \hline
\end{tabular}
\end{table}

\section{Methods} \label{section:Methods}

\subsection{Arps Forecasting Model}
To measure how accurate the forecasts of the suggested DNN models are, this paper compares the DNN models to the Arps decline curve model that is widely used by the industry and currently used by EIA. The Arps model was first introduced in 1945 by Arps \cite{Arps} and it empirically forecasts the decline in production of a specific well using historical data fit into a three-parameter model \citep{Okuszko_2008} as shown in Equation \ref{eq:1}:
\begin{center}
\begin{equation} \label{eq:1}
Qt = \frac{Qi}{\left( 1 + b * Di * t\right)^ {1/b}}
\end{equation}
\end{center}

\noindent where $Qt$ refers to the production at month $t$, $Qi$ is the production rate at time 0, $b$ is the line curvature degree, $t$ is the months in production and $Di$ is the initial decline rate. The Statistical and Analytical Agency in the US Department of Energy  \citep{EIA} provides county-specific parameters by averaging the production of all wells in the county and then fitting the average production to the decline curve using Equation \ref{eq:1}.\\
However, the Arps model uses only $Qi$ (first month of production) as input to generate predictions, while the DNN utilizes the first 4,6,8 or 10 months of production data. In order to provide a fair comparison between the two models, the Levenberg–Marquardt algorithm \citep{Leven,Marq,lmfit} (a non-linear least square fit) is used to find optimal $b$, $Qi$ and $Di$ parameters that give the best fit between the Arps model and the input data (first few months of production) as this approach is usually followed to utilize the first known months of actual production data. Simply put, for each well, the actual first 4,6,8 or 10 months of production data will be used to tune the Arps parameters so it would generate a better forecast.
This procedure provides a fair comparison between the DNN and the Arps model as it allows both models to utilize exactly the same data given as input.
\subsection{Deep Neural Network Forecasting} \label{Deep Neural Network Forecasting}
\begin{figure}
    \centering
    \includegraphics[width=14cm]{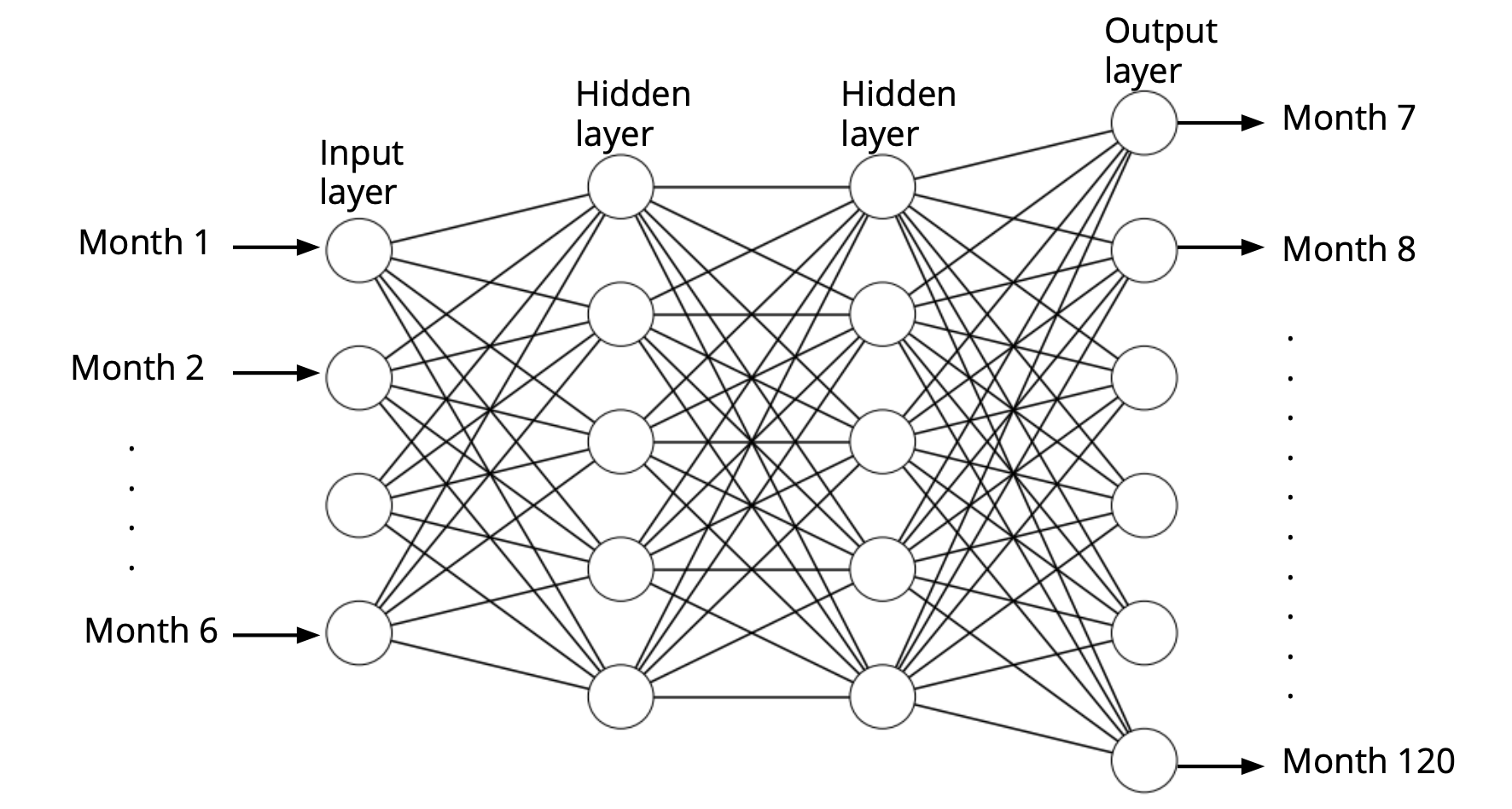}
    \caption{Layers used by the DNN model, which consists of an input layer with 30 neurons, two hidden layers
 with 35 and 50 neurons, and an output layer with a number of neurons equal  to the desired number
of output months. The first three layers are followed by a dropout layer with a value of 0.1 to reduce training data overfitting.}
    \label{fig:DNN_Sketch}
\end{figure}

A traditional supervised learning technique was used to train and test a DNN denoted as a mathematical function $\mathbbmss{F}$ to generate a predictive relationship between the input and output \citep{Hastle}, by mapping the input vector $X=\{x_1,x_2...,x_n\}$ into an output vector $Y=\{y_1,y_2,....,y_m\}$ where $n$ is the number of months used as an input to the DNN, $x_1$ is the first cumulative monthly production, $y_1$ denotes the first cumulative monthly production forecast, and $y_m$ denotes the last month production forecast.\\
The DNN model developed for this project is composed of a four-layer deep sequential neural network with an input dense layer of 30 neurons, two hidden dense layers with 35 and 50 neurons, and an output layer with a number of neurons equal to the desired number of output months as shown in Figure \ref{fig:DNN_Sketch}.
All the neurons in the network used a rectified linear unit (ReLU) as an activation function. In addition, three dropout layers with a value of 0.1 were used. The dropout layers were added after each of the first three layers to reduce the effects of overfitting \citep{dropout}.
The model was trained with the Adam stochastic optimizer \citep{Kingma} minimizing the mean absolute error (MAE) of the loss between the predictions and the labels. This model configuration was chosen as it resulted in the least testing error after extensive testing for all of the DNN parameters including all available activation functions and optimizers. In addition, the input and output data points (number of months) of the DNN are flexible and can be chosen during training time in order to see how the DNN will behave given a certain number of input months.\\
\subsection{Transfer Learning Models} \label{Transfer Learning Models}
\begin{figure}
    \centering
    \includegraphics[width=14cm]{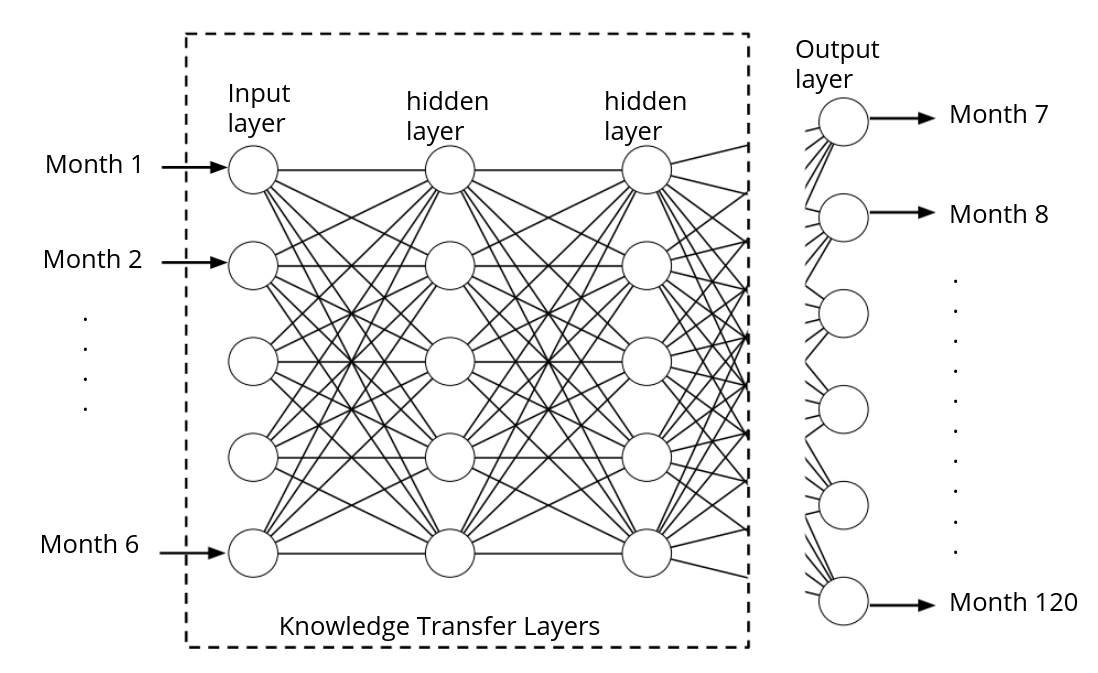}
    \caption{ In this paper, transfer learning is implemented by extracting the knowledge transfer layer from the source model $\mathbbmss{F}_S$ and using it along with a new untrained output layer to develop a target model $\mathbbmss{F}_T$. While training the transfer learning target model $\mathbbmss{F}_T$, the knowledge transfer layers must not be trained to preserve the knowledge that they have from training the source model $\mathbbmss{F}_S$.}
    \label{fig:Transfer_Learning_Implementation}
\end{figure}

DNNs are powerful tools and have been utilized heavily to tackle research problems, but one of their major drawbacks is that DNNs need an abundance of data to generate accurate predictions and in shale gas forecasting that can be a problem especially for counties with limited data. Transfer learning, which is a machine learning technique that can effectively use the knowledge from a pre-trained model to make predictions on a new set of data from a related problem \citep{Zhuang}, can be utilized to overcome the limitation of the available data. Transfer learning can be very useful in the case where the data set from the new problem is not large enough to properly train a DNN.\\ 
The formal definition of transfer learning in the context of machine learning as defined by \citet{YangBook} is: "Given a source domain $D_S$ and learning task $T_S$ , a target domain $D_T$ and learning task
$T_T$ , transfer learning aims to help improve the learning of the
target predictive function $\mathbbmss{F}_T$ in $D_T$ using the knowledge in
$D_S$ and $T_S$ , where $D_S \neq D_T$ , or $T_S \neq T_T$".
The goal is to use the knowledge stored in $\mathbbmss{F}_S$ acquired from the pre-trained model that had enough data for proper training and testing, to assist the new models' predictive function $\mathbbmss{F}_T$ in making accurate predictions despite data scarcity \citep{YangSurvey} .\\
For example, a marine biology research team wants to develop a DNN that can classify images of sharks and dolphins but does not have enough data (images) to train the DNN to generate accurate predictions. The team can use transfer learning by utilizing the VGG16 \citep{VGG} model as a source model $\mathbbmss{F}_S$ to help increase the accuracy of the predictions in their own target model $\mathbbmss{F}_T$. The VGG16 model is a convolutional DNN model that achieves more than 90\% accuracy in ImageNet \citep{ImageNet} which contains more than 10 million labeled images.
Although the VGG16 may not necessarily have been developed to classify images of sharks and dolphins, utilizing its ability of feature extraction \citep{brownlee3} can help generate a more accurate classifier for the sharks and dolphins pictures.\\
In this research paper, transfer learning is implemented by taking the pre-trained model $\mathbbmss{F}_S$ and removing its output layer that makes the predictions and only keeping prior layers (knowledge transfer layers) as shown in Figure \ref{fig:Transfer_Learning_Implementation} then designing the new model $\mathbbmss{F}_T$ by taking the knowledge transfer layers and adding a new untrained output layer to be trained only on the new data. However, it is very important that the knowledge transfer layers are not trained while training $\mathbbmss{F}_T$ to preserve the knowledge that they have from training the source model $\mathbbmss{F}_S$.
\noindent In this paper, transfer learning will be used to develop county-specific DNN models in order to compare the results to the Arps model benchmark since the Arps models are county-specific models as described by EIA \citep{EIA}. Transfer learning will be used to overcome the need for a large data set to properly train a DNN since most of these counties do not have enough samples (wells).\\
These county-specific DNN models are developed in the following manner, for each specific county in the state's data set a source model $\mathbbmss{F}_S$ was developed by training it on all of the available data except for that specific county. Then, using transfer learning, a target model $\mathbbmss{F}_T$ was trained and tested only on that county's data.
\subsection{Training and Testing Procedure}
For the DNN forecasting models, 75\% of the data is used for training while the remaining 25\% is used for testing. Due to the fact that the testing set is only used when the training of the model is complete, a small portion (10\%) of the training data is reserved for validation to monitor the error of the model during training where it helps the training algorithm perform better by providing it feedback to tune its hyperparameters before the learning process is finished \citep{Goodfellow_2016}. The DNN models are initially set to be trained for 200 epochs where in each single epoch the DNN model goes over the entire training data set and updates its weight accordingly using the back propagation algorithm \citep{nielsen_2020}. However, more training epochs than needed can cause overfitting because the optimization algorithm will over-optimize the weights of the model based only on the training data. This results in a model that does not generalize well and will produce a high error during testing \citep{Bishop}.
For this reason, an auto-stopping mechanism is used to monitor the validation score and if the validation score has not improved in the last 10 epochs then the training will automatically be stopped and the model with least error on the validation set will be considered as the forecasting model. The auto-stopping mechanism would prevent the model from overfitting or underfitting the data, save computing power and reduce unnecessary training time \citep{brownlee2}.

\subsection{Reproducibility and Verification}
\begin{figure}
    \centering
    \includegraphics[width=13cm]{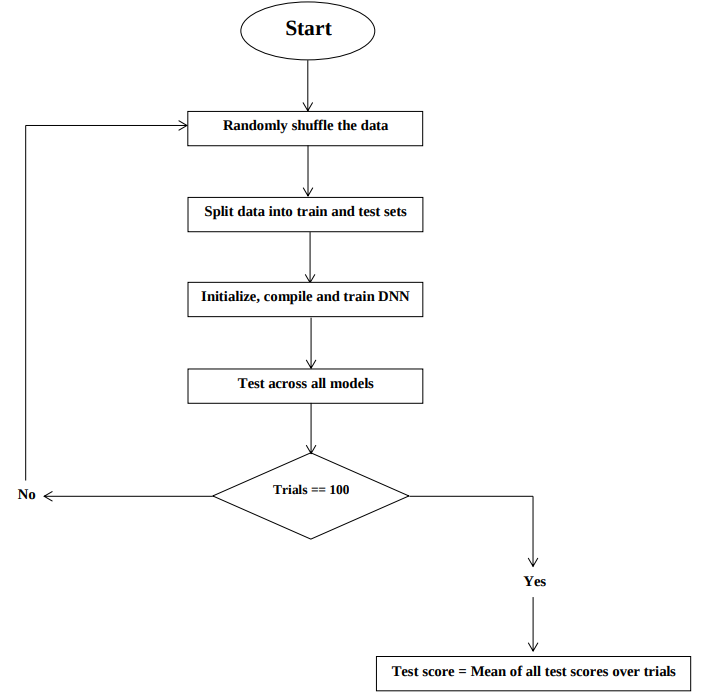}
    \caption{A flowchart showing the process of comparing the two forecasting models over the course of 100 trials to address both the reproducibility concerns caused by the randomness in the DNN and to ensure a fair comparison between the models of interest across multiple test sets.}
    \label{fig:rep_flowchart}
\end{figure}

In DNN and in machine learning in general, one of the main challenges is reproducibility \citep{Huston} which is difficult to achieve due to the random initialization of the weights prior training in addition to many other factors such as using a stochastic optimization algorithm. The problem complicates the efforts of quantifying the improvement of the suggested enhanced forecasting models due to the fact that each time a model is trained and tested it generates slightly different results.
To make sure that the improvement in the forecasts was not caused by such randomness, the process of initializing, compiling, training, and testing of each model were repeated 100 times and the average of those 100 runs was reported as the final result.\\
Furthermore, to take into account the possibility that the DNN models outperform the Arps model only on a specific testing set (i.e, Arps may perform similar or better if the testing set is changed), at each run before the data is split into training and testing sets the data is randomly shuffled. However, it is important to mention that although the data is shuffled at the beginning of each run, the measurement of the forecasting error in terms of MAE across all models is done on exactly the same testing data set for that run. Figure \ref{fig:rep_flowchart} explains the flowchart of the comparison process between the models of interest.

 \section{Results} \label{ResultsSection}
 \subsection{Deep Neural Networks}
\begin{figure}
    \centering
    \includegraphics[width=14cm]{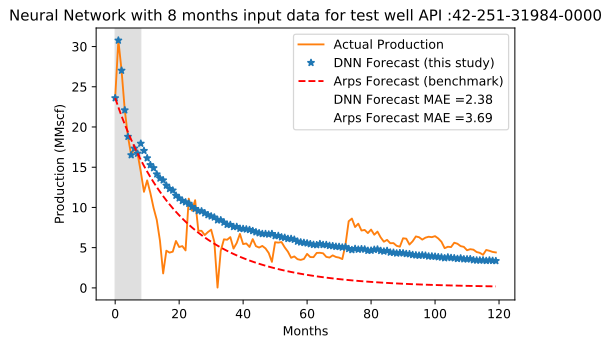}
    \caption{A sample test well from the Barnett shale shows the DNN model and the Arps model forecasting compared to the actual production data where 8 months of input data is used to forecast the production. The plot shows that with enough data, the DNN model can generate a more accurate production forecast than the Arps model. The gray area on the left of the plot indicates the data points used as input to the forecasting models while the rest of the plot shows the production forecast.}
    \label{fig:Barnett_Sample_well}
\end{figure}

\begin{figure}
    \centering

\includegraphics[width=12cm]{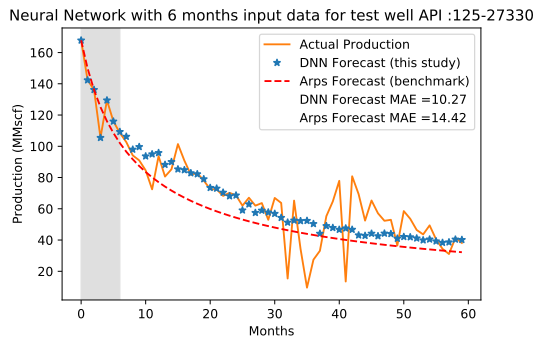}
    \caption{A sample test well from the Marcellus shale shows the DNN model and the Arps model forecasts compared to the actual production data where 6 months of input data is used to forecast the production. The plot shows that with enough data, the DNN model can generate a more accurate production forecast than the Arps model. The gray area on the left of the plot indicates the data points used as input to the forecasting models while the rest of the plot shows the production forecast.}
    \label{fig:Marcellus_Model_Sample_Forecast}
\end{figure}

As described in Section \ref{Deep Neural Network Forecasting}, two DNN models were trained and tested where the first DNN model was developed on the Barnett data set and the other one on the Marcellus data set.
A sample well production forecast is shown in Figure \ref{fig:Barnett_Sample_well} and Figure \ref{fig:Marcellus_Model_Sample_Forecast} where the grey area on the left of the plot indicates the data points (number of input months) used as an input to generate the forecast while the rest of the plot shows the models' predictions. To measure the accuracy of the forecasting models, each model is compared to the actual production in terms of MAE. \\
\noindent Both figures show that DNN can generate more accurate production forecasts than the Arps model. The DNN models were able to achieve more than 35\% error reduction on the Barnett sample well and more than 28\% error reduction on the Marcellus sample well. These results show that DNN models have the ability to map input data into useful output and generate accurate production forecasts. However, the two DNN models are statewide-level models and the Arps models are county-specific models so in order to fairly compare the results across all the wells in the testing set we need to compare DNN county-specific models to the current Arps county-specific model benchmark.

\subsection{Transfer Learning}
As described in Section \ref{Transfer Learning Models}, this research paper employed the technique of transfer learning to develop a better shale gas production forecast than the Arps benchmark model. Tables 3 and 4 show that in both data sets most counties do not have enough samples to properly train a DNN.

\begin{table}[H]
\centering
\small
\begin{tabular}{|p{1cm}|p{0.75cm}|p{0.75cm}|p{0.75cm}|p{0.75cm}|p{0.65cm}|p{0.5cm}|p{0.75cm}|p{0.5cm}|p{1.5cm}|p{0.5cm}|p{1.25cm}|}
 \hline
 \multicolumn{12}{|c|}{Table 3: Number of wells per county in the Barnett data set} \\
 \hline
Johnson &Tarrant &Denton  &Parker & Wise&Hood &Hill& Erath& Jack &Palo Pinto  & Ellis &Somervell \\
 \hline
1372 & 1050 &620 &469& 425&272&131&36&24&16&12&12\\

 \hline
\end{tabular}
\end{table}

\begin{table}[H]
\centering
\small
\begin{tabular}{|p{2.5cm}|p{2.5cm}|p{2.5cm}|p{2.5cm}|p{2.5cm}|}
 \hline
 \multicolumn{5}{|c|}{Table 4: Number of wells per county in the Marcellus data set} \\
 \hline
Susquehanna &Greene &Wyoming  &Washington & Westmoreland \\
 \hline
658 & 535 &94 &703& 182\\

 \hline
\end{tabular}
\label{tab:Marcellus_Counties_Wells}
\end{table}

\begin{figure}
    \centering
    \includegraphics[width=14cm]{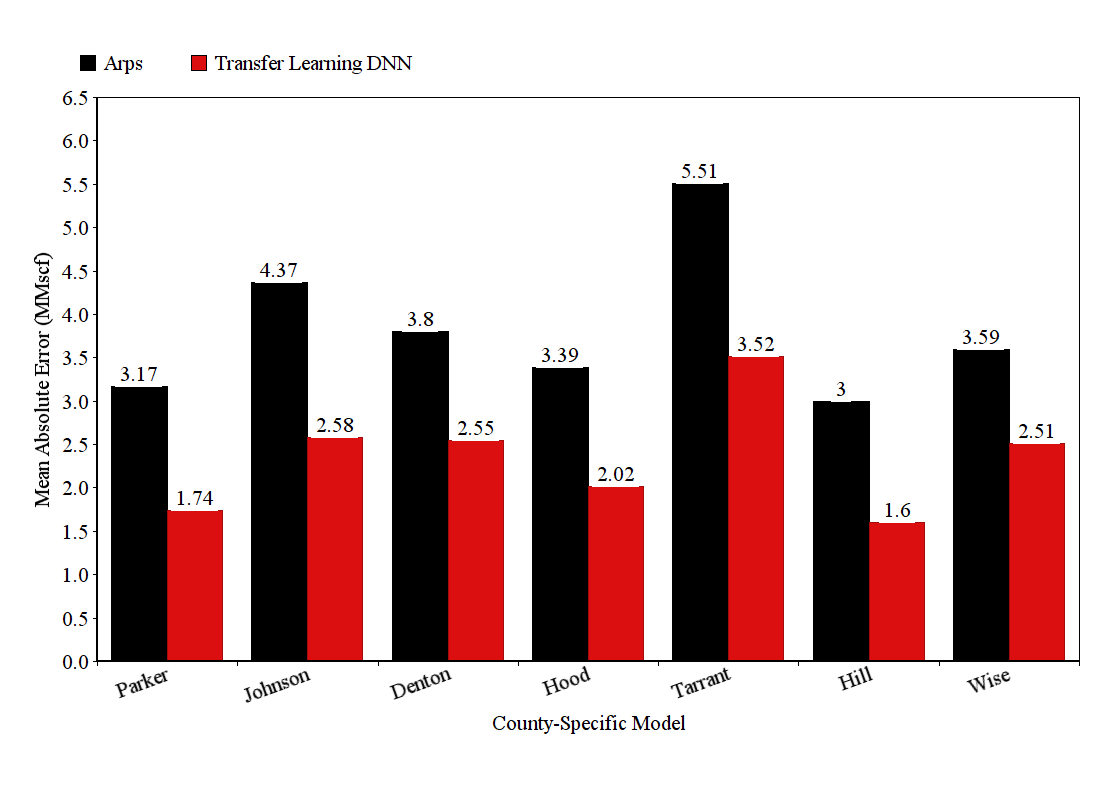}
    \caption{The new suggested county-specific DNN models achieved an error reduction between 47\% in Hill County and 30\% in Wise County compared to the Arps model benchmark when 6 months of input data were used to generate both forecasts.}
    \label{fig:TL_Barnett}
\end{figure}

\begin{figure}
    \centering
    \includegraphics[width=14cm]{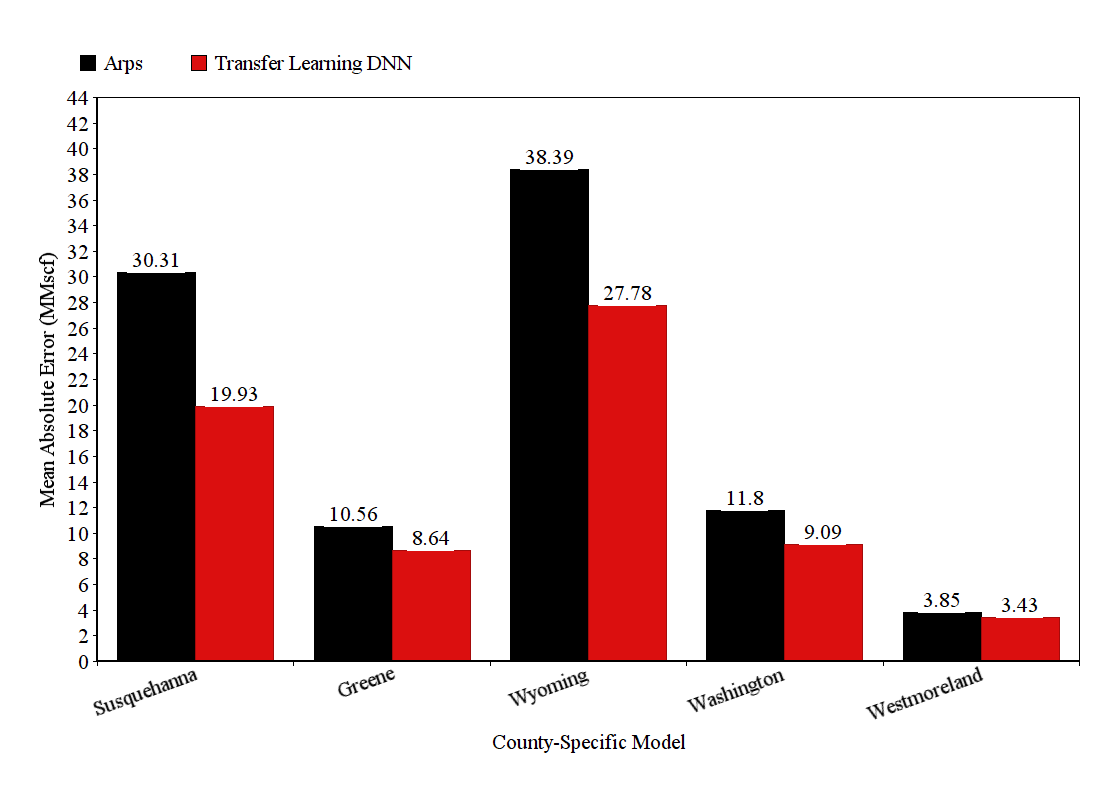}
    \caption{The new suggested county-specific DNN models achieved an error reduction between 34\% in Susquehanna County and 11\% in Westmoreland County compared to the Arps model benchmark when 6 months of input data were used to generate both forecasts.}
    \label{fig:TL_Marcellus}
\end{figure}

\begin{figure}
    \centering
    \includegraphics[width=14cm]{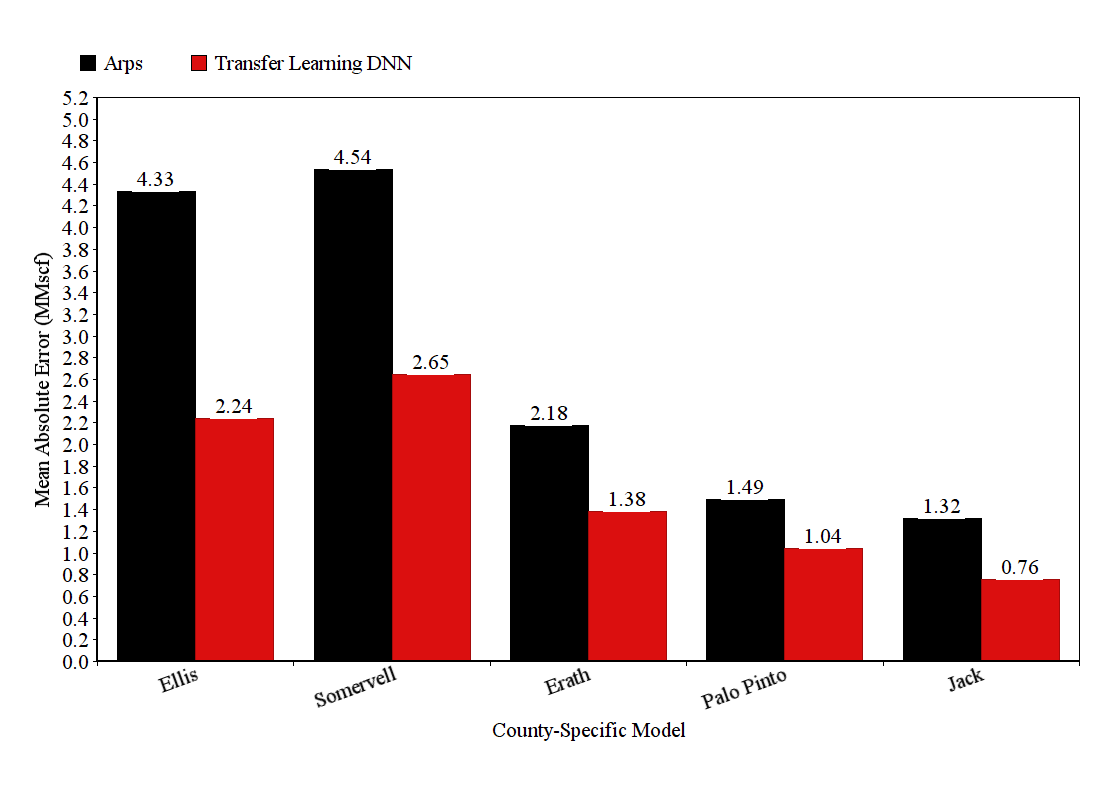}
    \caption{County-specific DNN models were able to achieve a significant error reduction compared to the Arps model ranging between 30\% in Palo Pinto County and 48\% in Ellis County  despite the fact that the DNN model was not trained on that specific county data and the whole county's data was used for testing. The use of transfer learning enabled the DNN models to generate accurate forecast despite data scarcity in these counties. The source model $\mathbbmss{F}_S$ was trained on all other counties in the state except for the county of interest then all of this county's data was used for testing. }
    \label{fig:TL_Data_scarcity}
\end{figure}

\noindent After following the procedure described in the Methodology Section \ref{Transfer Learning Models} and applying it on both the Barnett and the Marcellus data sets, the DNN transfer learning models were able to reduce the error significantly. Figure \ref{fig:TL_Barnett} shows that when this method is used with 6 months of input data on the Barnett shale data set, DNN transfer learning models were able to reduce the error between 47\% in Hill County and 30\% in Wise County when compared to the Arps benchmark model. Similar results as shown in Figure \ref{fig:TL_Marcellus} is obtained from applying the same method on the Marcellus shale data set where an error reduction between 34\% in Susquehanna County and 11\% in Westmoreland County is achieved compared to Arps.
\noindent However, in the Barnett data set the counties of Somervell, Ellis, Palo Pinto, Jack and Erath do not even have enough data to train a transfer learning model where the number of samples is very small to even train the last layer and develop a good predictive model. For this reason, the source model $\mathbbmss{F}_S$ which was trained on all other counties in that data set except for the county of interest was imported and used as is without any training where all that county's data will be used for testing. The model is then used to forecast the production in the county of interest. Figure \ref{fig:TL_Data_scarcity} shows that when 6 months of input data is used, an error reduction between 48\% in Ellis County and 30\% in Palo Pinto County is achieved compared to the Arps model.

\begin{figure}
    \centering
    \includegraphics[width=14cm]{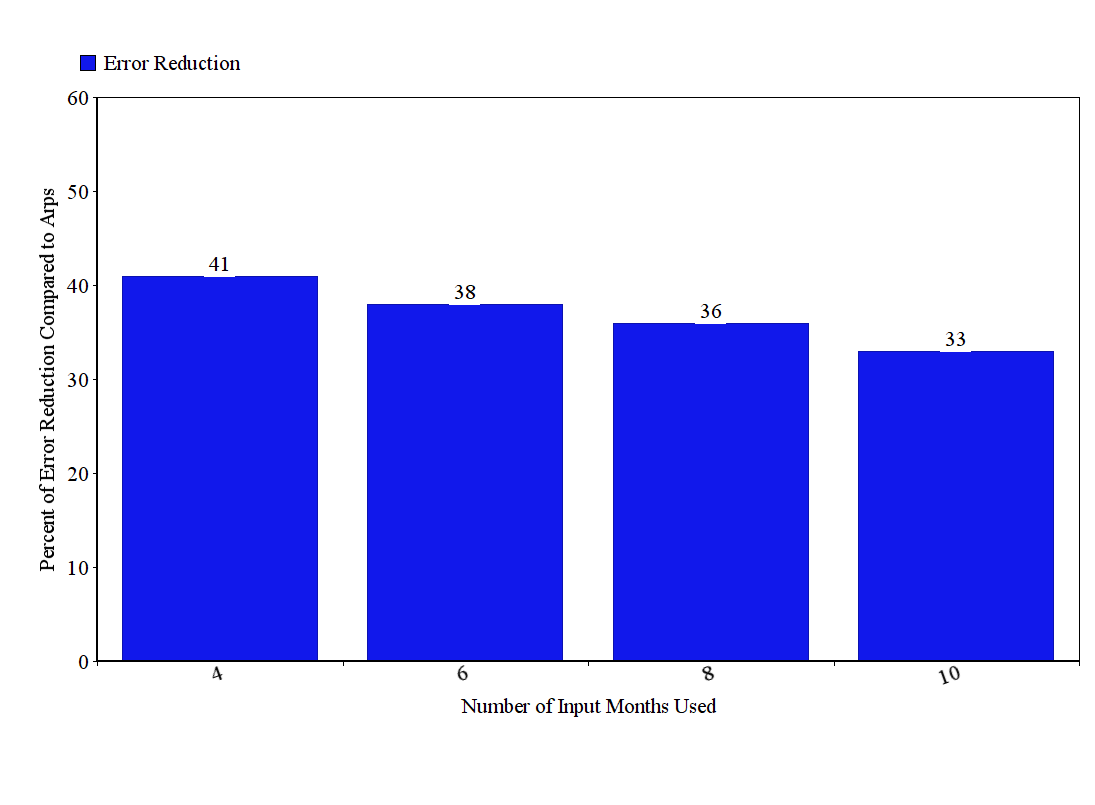}
    \caption{The error reduction achieved when using the county-specific DNN models compared to the Arps models on the whole Barnett data set. The error reduction is computed as the weighted average error reduction across all counties in the data set using Equation \ref{eq:2} when different values of input months is used.}
    \label{fig:TL_Barnett_Error_red}
\end{figure}

\begin{figure}
    \centering
    \includegraphics[width=14cm]{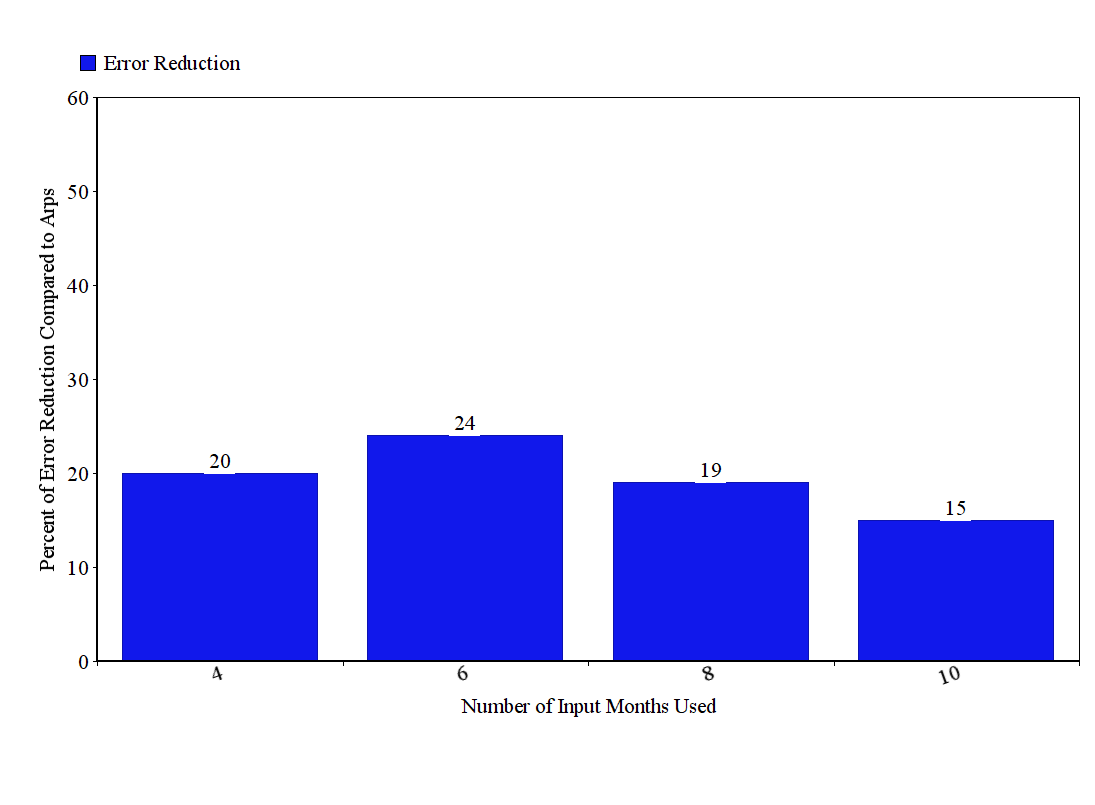}
    \caption{The error reduction achieved when using the county-specific DNN models compared to the Arps models on the whole Marcellus data set. The error reduction is computed as the weighted average error reduction across all counties in the data set using Equation \ref{eq:2} when different values of input months is used.}
    \label{fig:TL_Marcellus_Error_red}
\end{figure}

\noindent Figures \ref{fig:TL_Barnett_Error_red} and \ref{fig:TL_Marcellus_Error_red} show the error reduction compared to Arps when using county-specific DNN models across all counties for each state. The percent reduction in error shown in the figures represents the weighted average error reduction on the all of the testing sets across all counties as shown in Equation \ref{eq:2}:
\begin{center}
\begin{equation} \label{eq:2}
\text{Overall Error Reduction} = \frac{\sum_{i} {\text {Error Reduction in County(i)} \times \text{Number of test wells in County(i)}}}{\text {Number of test wells in all counties} }
\end{equation}
\end{center}
The figures show that the DNN approach consistently outperformed the Arps model across all counties and input months used as it averaged an error reduction on the Barnett shale data set between 41\% when 4 months of input data is used to 33\% when 10 months of input data is used and between 24\% and 15\% on the Marcellus shale data set.

\noindent The county-specific DNN models performed better on the Barnett data set than they did on the Marcellus data set due to the fact that production history available is longer allowing the DNN to learn better with more data. This shows that the county-specific DNN models can be improved further as more production history becomes available. The web pages of \citet{TX_website} and \citet{PA_website} publish new production data frequently giving the new approach an advantage over the Arps models which tend to overestimate the production in the middle and late life of a well.

\section{Discussion} \label{Discussion}
\begin{figure}
    \centering
    \includegraphics[width=14cm]{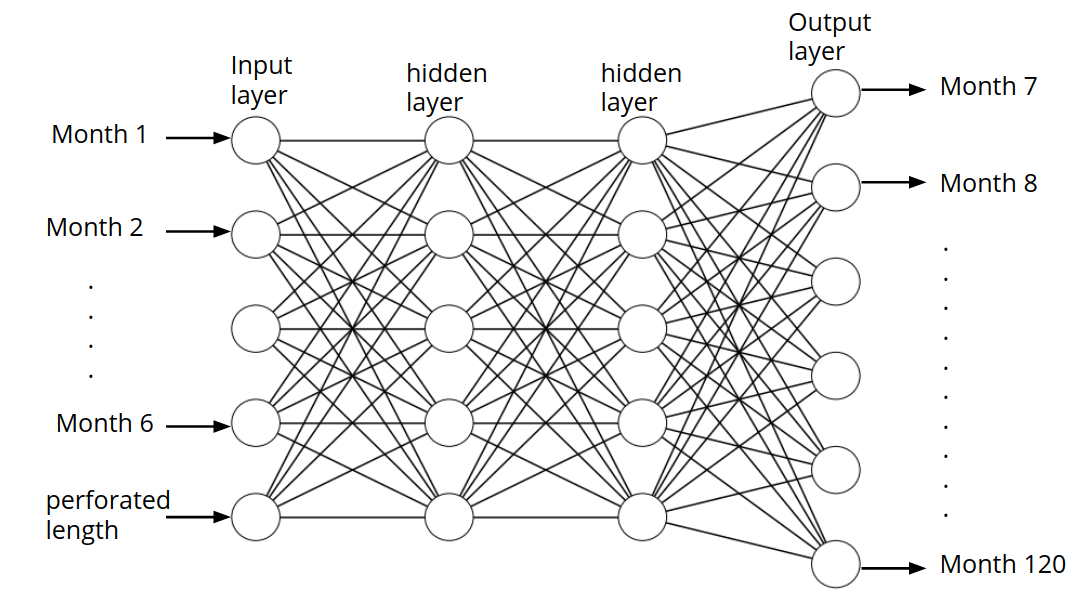}
    \caption{Adding well-specific physical properties such as perforated length, fracking fluid volume or sand volume may introduce further improvement to the DNN forecasting model as the DNN may be able to learn the relationship between such parameters and the rate of decline in production.}
    \label{fig:perf_as_inpt}
\end{figure}

As the results show in Section \ref{ResultsSection}, DNN models can significantly reduce the error in forecasting compared to the Arps models.
The use of transfer learning to develop county-specific DNN models generated a more accurate production forecast and provided the chance to develop DNN models for counties with very limited numbers of samples. 
One of the main reasons behind the improvement introduced by the DNN models is they have the ability to fit the data on models with up to 800 million parameters \citep{Mahajan} giving them the power to generate models that can map very complex data. The DNN developed for this research paper uses more than 5000 parameters to generate its forecast while the Arps model only uses three and unlike Arps parameters which are provided by EIA as described in the Methods Section \ref{section:Methods}, the DNN tunes its parameters directly from the data through the training process in order generate a model that generalizes well on out-of-sample test data. The advancement in computing power along with efficient machine learning libraries and optimizers gave the chance to tune those 5000 parameters on thousands of samples in a few minutes on an average laptop.\\
As production data continue to grow due to numerous shale gas wells currently in production, the DNN models can be further improved by more data as they scale very well with data. Unlike the Arps model where the ill-posedness limits its ability to improve with more data, as the more the data becomes available the more the fluctuation in production can occur between different wells leading to non-unique parameter estimates. This study showed using transfer learning to build county-specific DNN models offer a significant error reduction as the county-specific DNN model allowed the DNN to generate a more accurate forecast due to the fact the model can utilize the knowledge gained from other nearby counties then fine-tune its output (forecasting) layer specifically for the county of interest. Besides increasing the accuracy of the forecasts, transfer learning also helps overcome the data scarcity problem faced when a certain county or area of interest might not have enough data to properly train a DNN.\\
Transfer learning introduced a way of utilizing data from other counties' aggregate DNN model into a county-specific model despite the fact that two DNN models may have different domains $D_T$ and $D_S$ where the domain in this case refers to the geological formation. Although the two models have different domains, there is still similar features in the pattern of production decline and transfer learning allowed the target model $\mathbbmss{F}_T$ to improve its forecasting based on those similar patterns extracted by the source model $\mathbbmss{F}_S$. Even in the counties of Somervell, Ellis, Palo Pinto, Jack and Erath where data is not even enough to train a transfer learning model, the model that was trained only on adjacent counties' data was able to achieve significant error reduction thanks to the ability of DNN pattern extraction. However, there is still uncertainty on how much the change in geological properties between the two domains $D_T$ and $D_S$ is affecting the results. The uncertainty is caused by the heterogeneity of the geology which can be heterogeneous even at a small scale \citep{Justin2017}. Unfortunately, the data sets available do not contain geological information data which makes it difficult to investigate such impact on the results.\\ 
In addition to the uncertainty emerging from the variation in geology, both the Arps and the DNN models have a major uncertainty regarding the effect of the change in technology and how that might affect the forecasting on a new well especially because these models are data driven and do not take into account the physical properties of each well. Developing a DNN model that utilizes physical properties to enhance forecasting may introduce further improvement to the forecasting. Similar to \citet{Sun_2018} where they used well head pressure as input, a good assumption can be made that further improvement can be achieved by incorporating a well-specific physical property such as perforated length, fracking fluid volume or sand volume, etc., as an input to the DNN as shown in Figure \ref{fig:perf_as_inpt} since these parameters have an impact on the decline in production and the DNN may be able to learn the relationship between these parameters and the production decline. For example, in general a longer perforated length in a certain well leads to more production \citep{Xia} where the DNN may be able to learn such a relationship between the perforated length and the rate of decline to enhance its forecasting. Unfortunately, due to data limitation on these properties such hypothesis was not tested.
\\ \noindent The described DNN approach has a disadvantage compared to Arps models as DNN forecasting requires multiple models depending on the number of input months used to generate forecasts due to the fact the DNN will tune its weights during training specifically to map the input data of a certain length to the output data. For example, if a well has seven months of data and we want to forecast the next ten years of production then only the previously trained seven months DNN model can be used. This causes the need for developing multiple DNN models for each field or county of interest in order to get a flexibility in choosing the number of input months available to generate the forecast. Although the process of building DNN models requires more work to develop compared to the Arps forecasting models, the DNN training process to generate the models can easily be automated and is only needed once at the beginning then these DNN models can be used efficiently to generate the forecasts. 
\section{Conclusion}
In this paper the results show that DNNs can generate much better forecasts than the Arps decline curves by using transfer learning to develop county-specific DNN models fine-tuned to the county of interest as they were able to reduce the forecasting error up to 47\% compared to the Arps decline curve model. The new suggested approach improves the current forecasting techniques used which is crucial for decision making and calculating returns on investment. 
Moreover, the results also show that the need for a large amount of data to produce accurate shale gas forecasts is no longer a hurdle for machine learning forecasting models as transfer learning can be used to overcome the data scarcity arising from a limited number of samples by transferring the knowledge gained from the other counties in the data set.\\
By offering a mechanism that allows the use of data from a specific area of interest as well as other nearby areas, this approach also offers the chance to easily improve the forecasting models further with more data as more production data will be available in the future. A realistic goal would be to gather all the production data from all the shale formations in the US to build a large model for each formation to serve as a source function $\mathbbmss{F}_S$ to transfer the knowledge it holds into county-specific DNN models that can produce much better forecasting results than the current widely used Arps model. 

\section*{Acknowledgement}
Omar Alolayan was supported by a graduate fellowship from Saudi Aramco.
\FloatBarrier
\bibliographystyle{unsrtnat}


\end{document}